\documentclass{article}
\makeatletter
\def\blfootnote{\gdef\@thefnmark{}\@footnotetext}
\makeatother
\usepackage{amsmath,graphicx,hyperref}
\usepackage{overpic}
\usepackage{adjustbox}
\usepackage{siunitx}
\usepackage{tabularx}
\usepackage[table,xcdraw]{xcolor}
\usepackage{subcaption}
\usepackage{amsfonts}
\usepackage[ruled,vlined]{algorithm2e}
\usepackage{algorithmic}
\usepackage{mathtools}

\begin{document}
\title{Disentanglement of Sources in a \\ Multi-Stream Variational Autoencoder}

\author{Veranika Boukun$^{\star}$ 
\qquad Jörg Lücke$^{\star\dagger}$ 
\\
$^{\star}$Machine Learning Lab, Department of Medical Physics and Acoustics, \\Carl von Ossietzky Universität Oldenburg, Germany \\
$^{\dagger}$Research Group AI, Department of Computer Science,\\ University of Innsbruck, Austria \\ veranika.boukun@uol.de, joerg.luecke@uibk.ac.at}
\date{\phantom{nothing}}

\maketitle

\begin{abstract}
Variational autoencoders (VAEs) are among leading approa\-ches to address the problem of learning disentangled representations. Typically a single VAE is used and disentangled representations are sought within its single continuous latent space. In this paper, we propose and provide a proof of concept for a novel Multi-Stream Variational Autoencoder (MS-VAE) that achieves disentanglement of sources by combining discrete and continuous latents. The discrete latents are used in an explicit source combination model, that superimposes a set of sources as part of the MS-VAE decoder. We formally define the MS-VAE approach, derive its inference and learning equations, and numerically investigate its principled functionality. The MS-VAE model is very flexible and can be trained using little supervision (we use fully unsupervised learning after pretraining with some labels).  
In our numerical experiments, we explored the ability of the MS-VAE approach in separating both superimposed hand-written digits as well as sound sources. For the former task we used superimposed MNIST digits (an increasingly common benchmark). For sound separation, our experiments focused on the task of speaker diarization in a recording conversation between two speakers. In all cases, we observe a clear separation of sources and competitive performance after training. For digit superpositions, performance is particularly competitive in complex mixtures (e.g., three and four digits). For the speaker diarization task, we observe an especially low rate of missed speakers and a more precise speaker attribution. Numerical experiments confirm the flexibility of the approach across varying amounts of supervision, and we observed high performance, e.g., when using just 10\% of the labels for pretraining. 
\end{abstract}

\section{Introduction and Related Work}
\label{sec:intro}
Disentanglement is one of the sought-after properties in recent machine learning (ML) research, and, though definitions vary, the commonly followed definition from \cite{bengio2013representation} states that for a disentangled representation: (1) variations in high-dimensional input data are explained by semantically meaningful low-dimensional latent codes that reflect distinct factors of variation; (2) changes in a single underlying factor affect only one component of the learned representation; and (3) latent variables are statistically independent. Disentangled Representation Learning (DRL) leverages machine learning models to identify and separate meaningful latent variables within observed data \cite{wang2024disentangled}. Variational autoencoders (VAEs)\cite{Kingma2014} are among the most popular DRL approaches, as standard VAEs promote uncorrelated, though not always fully disentangled, representations \cite{RolinekEtAl2019}.

\begin{figure}[t]
    \centering
    \includegraphics[width=\linewidth]{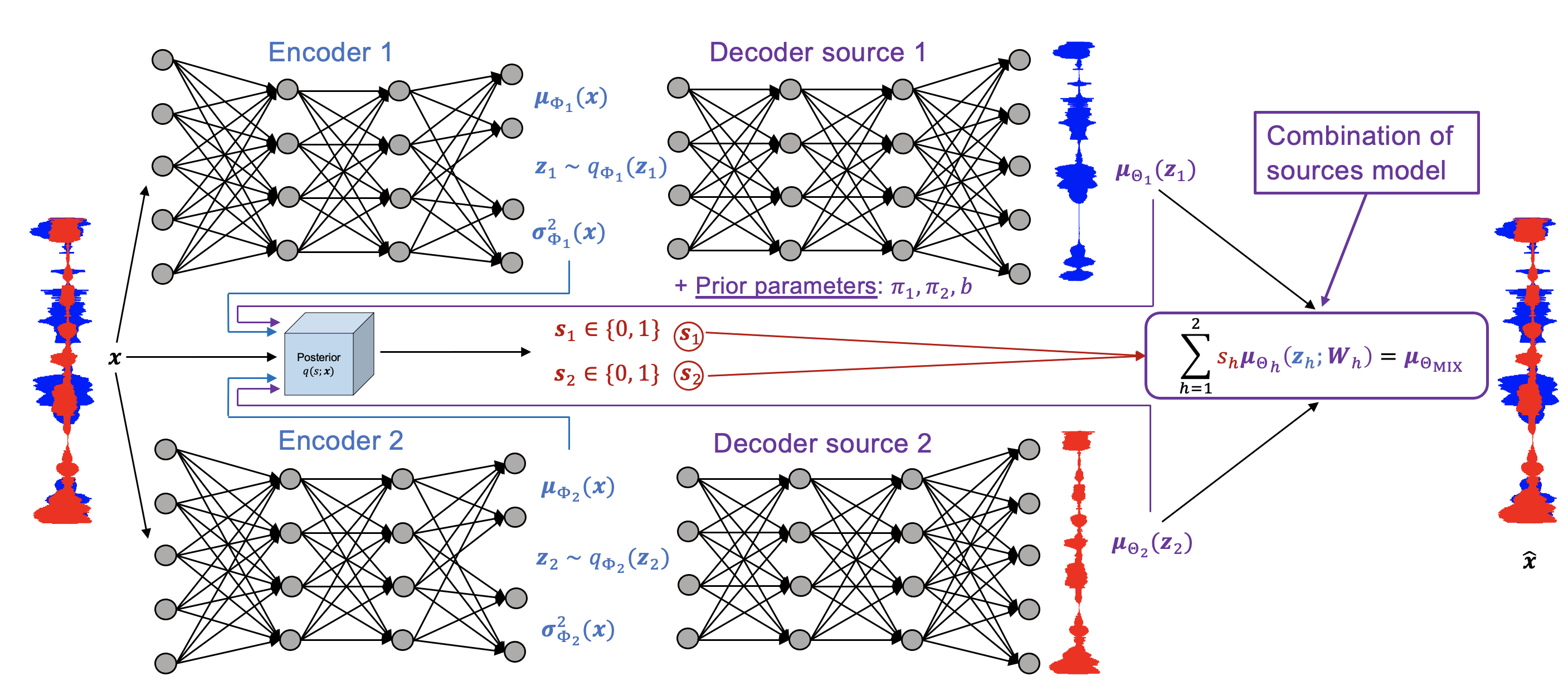}
    \caption{Schematic representation of MS-VAE for 2 sources ($H=2$).}
    \label{fig:ms-vae-diagram}
\end{figure}

Among others, one difficulty in achieving disentanglement is the representation of discrete properties alongside continuous variation in object attributes or signal properties. Examples of such discrete factors might include object's presence or absence in an image or speaker activities in an acoustic scene. Typically, VAE-based disentanglement approaches rely only on a single continuous latent space, in which one seeks disentangled features (see section on \hyperref[sec:related_work]{Related Work}), however, much real-world data (including images and acoustic signals) naturally exhibit both discrete and continuous variation. Modeling with discrete latents can offer greater interpretability and improved disentanglement. However, in practice, the direct use of discrete variables is often avoided because they introduce optimization challenges (e.g., gradients cannot be propagated through non-continuous encodings). Instead, stochastic gradient estimators such as REINFORCE \cite{williams1992reinforce} with variance control techniques, or continuous relaxations like the Gumbel Softmax \cite{jang2016categorical,maddison2017concrete} are commonly used in approaches that aim to incorporate discrete factors of variation. Direct discrete latent optimization is, however, possible in practice, as has been shown in, e.g., \cite{drefs2022evolutionary,drefs2023direct}.

In this paper, we propose a variational autoencoder framework called \textit{multi-stream VAE} (MS-VAE), which explicitly combines discrete and continuous latent representations using an explicit \textit{the combination of sources} model. Our approach is motivated by the structured nature of sources: discrete latent variables model the activity (i.e., presence or absence) of sources in any given observation, while continuous latent variables capture the signal properties, which are required, e.g., for signal reconstruction. In contrast to the approaches discussed in \hyperref[sec:related_work]{Related Work}, we here aim to disentangle sources using discrete latents which are optimized directly without continuous relaxations. Our approach employs a proper, unmodified variational lower bound without additional weightings or heuristic loss terms. Crucially, source combinations are handled explicitly using a linear combination model. The model is well-suited for many scenarios (including waveform mixtures), and it enables interpretable separations of sources. Each source is modeled by an independent, standard VAEs. The resulting MS-VAE approach is thus a hybrid framework combining discrete and continuous latents with different roles.

\subsection{Related Work.} \label{sec:related_work} Numerous extensions and modifications of VAEs have been proposed to promote disentanglement in their latent spaces. $\beta$-VAE \cite{higgins2017beta} approaches use weight coefficients to enhance disentanglement at the expense of reconstruction quality. \cite{burgess2018understanding} proposes to extend $\beta$-VAE to use an information bottleneck formulation which improves the trade-off of disentanglement and reconstruction quality. Disentangled Inferred Prior (DIP)-VAE-I and DIP-VAE-II \cite{kumar2017variational} incorporated and consistency terms and regularization on the expectation of the approximate posterior to promote disentanglement. The Factor VAE approach \cite{kim2018disentangling} imposes independence constraints via total correlation. $\beta$-Total Correlation VAE ($\beta$-TCVAE) \cite{chen2018isolating} uses decomposition of the KL-divergence to three terms (mutual information, total correlation and dimension-wise KL divergence) which are weighted separately to improve disentanglement performance. For Relevance Factor VAE (RF-VAE) \cite{kim2019relevance}, relevance indicators were used to recognize meaningful factors of variation. \cite{ye2021deep} introduce a deep variational framework of Mixtures of VAEs (MVAE), where multiple encoders acting as mixture components are used to capture different variational features and a weighted regularization (Hilbert-Schmidt independence criterion) is added to the ELBO in order to promote the desired disentanglement of learned features. 

The aforementioned DLR approaches (expect \cite{ye2021deep}) focus on disentanglement in single continuous latent spaces. There are also VAE-based DRL methods that incorporate both continuous and a discrete variation factor (approximated by Gumbel Softmax priors) such as JointVAE \cite{dupont2018learning} and Multi-VAE \cite{xu2021multi} or using a categorical prior as in InfoGAN \cite{chen2016infogan}. We would like to highlight that, in contrast to the here proposed MS-VAE, the approaches listed above either do not treat discrete latent variables directly (i.e., by using continuous relaxations such as Gumbel Softmax in \cite{dupont2018learning,xu2021multi}), they add auxiliary loss terms and/or weightings to the ELBO objective as in \cite{burgess2018understanding,chen2018isolating,higgins2017beta,kim2018disentangling,kim2019relevance,kumar2017variational,ye2021deep}, or they have no properly defined variational lower bound as in \cite{chen2016infogan}).

\section{Proposed Method}
\label{sec:proposed_method}
Let us denote observed data by $\vec{x}^{\,(1:N)}\in \mathbb{R}^D$, use $s_{1:H}\in{}\{0,1\}^H$ for the discrete latents, and $\vec{z}_h\in \mathbb{R}^Z$ for the continuous latents (one $\vec{z}_h$ for each source $h$, and we abbreviate $\vec{z}_{1:H}$ by~$\vec{z}$). 
The MS-VAE generative model (decoder) $p_\Theta (\vec{x})$ is then given~by:
\begin{align}
    p_\Theta (\vec{s}) &= \textstyle\prod_{h=1}^{H}\pi_h^{s_h}(1- \pi_h)^{(1-s_h)}, 
    \label{eq:gen_model_1}\\
    p_\Theta (\vec{z}) &= \textstyle\prod_{h=1}^{H}\mathcal{N}\big(\vec{z}_h; \vec{0}, \mathbb{I}\big),
    \label{eq:gen_model_2}\\
    p_\Theta (\vec{x}|\vec{s},\vec{z}) &= \mathrm{Lap}\big(\vec{x}; \textstyle\sum_{h=1}^{H}s_h\vec{\mu}(\vec{z}_h; W_h), b), 
    \label{eq:gen_model_3}
\end{align}
where $\vec{\mu}(\vec{z}_h; W_h)$ is the decoder DNN for source $h$ with weights and biases given by $W_h$, $W_{1:H}$ are the parameters of all decoder DNNs. The set of all model parameters is thus given by $\Theta=(\vec{\pi},W_{1:H},b)$, where $\vec{\pi} = \pi_{1:H}\in[0,1]^H$ are the parameters for the Bernoulli prior, and $b$ is the noise level of the used Laplace distribution for the observed data. 
A crucial component of the MS-VAE is the explicit source combination model given in Eq.~\eqref{eq:gen_model_3}. Here all $H$ sources are explicitly combined using $\vec{s}$ and we will in the following use 
\begin{equation}
\textstyle\vec{\mu}_{\Theta_\mathrm{mix}} = \sum_{h=1}^{H}s_h\vec{\mu}(\vec{z}_h; W_h)
    \label{eq:comb_model}
\end{equation}
to abbreviate the combination model.

The optimization of MS-VAE model parameters is (as for a standard VAEs) based on an approximate maximum likelihood approach: we seek parameters $\Theta^*$, such that $\Theta^* = \underset{\Theta}{\mathrm{argmax}}\,\mathcal{L}(\Theta) = \underset{\Theta}{\mathrm{argmax}}\sum_{n}\log p_\Theta (\vec{x}^{\,(n)})$. 
Instead of directly optimizing the log-likelihood, we optimize a variational lower bound (ELBO). The ELBO is defined by the generative model Eqs.~\eqref{eq:gen_model_1}-\eqref{eq:gen_model_3} and by an approximation of the exact posterior $p_{\Theta}(\vec{s},\vec{z}\,|\,\vec{x})$. Here we use an approximation of the following from:
\begin{align}
    p_{\Theta}(\vec{s},\vec{z}\,|\,\vec{x}) \approx q_{\Phi}(\vec{s},\vec{z}; \vec{x})=q(\vec{s};\vec{z}, \vec{x}) \, q_{\Phi}(\vec{z}; \vec{x}).
    \label{eq:post_approx}
\end{align}
The used approximate posterior generalizes the original form of ELBO \cite{Kingma2014} and leads to closed-form expressions for all Kullback-Leibler divergences.
We assume the distributions $q_{\Phi_h}(\vec{z}_h; \vec{x})$ to be standard amortized variational distributions, i.e., 
\begin{align}
    q_{\Phi_h}(\vec{z}_h; \vec{x}^{(n)})={\cal N}\big(\vec{z}_h; \vec{\nu}_h(\vec{x}^{(n)}; \Phi_h),\mathrm{diag}(\vec{\tau}_h(\vec{x}^{(n)};{\Phi_h}))\big),
    \label{eq:encoder}
\end{align}
where we denote by $\vec{\nu}_h(\vec{x}; \Phi_h)$ the encoder DNN of source $h$ for the mean. The covariance matrix $\mathrm{diag}(\vec{\tau}_h(\vec{x};{\Phi_h}))$ is a diagonal matrix with entries $\tau_z(\vec{x};{\Phi_h})$, and the $Z$-dim vector $\vec{\tau}_h(\vec{x};{\Phi_h})$ is defined by the encoder DNN. We use $\Phi_h$ to denote all weights and biases for the encoder DNNs for source~$h$, and $\Phi_{1:H}$ are parameters of all encoder DNNs. Further we use the notation $q_{\Phi}(\vec{z}; \vec{x}^{(n)}) =\textstyle \prod_{h=1}^H  q_{\Phi_h}(\vec{z}_h; \vec{x}^{(n)})$ for all encoders (Mean field approximation).
Based on the definitions for encoders and decoders of the MS-VAE model, its ELBO objective can be canonically derived and is given by:
\begin{align}
  \mathcal{F}(\Phi, \Theta) &= \nonumber \dfrac{1} {N}\biggl(\textstyle\sum_{n}\int_{\vec{z}}\sum_{\vec{s}}{q(\vec{s};\vec{z}, \vec{x}^{(n)})}{q_{\Phi}(\vec{z}; \vec{x}^{(n)})}\big[ \log p_{\Theta}(\vec{x}^{(n)}|\vec{s},\vec{z}) + \log p_\Theta (\vec{s}) \\
  &- \log q (\vec{s};\vec{z}, \vec{x}^{(n)})\big]\mathrm{d}\vec{z} - \textstyle\sum_{n}\sum_{h}\mathrm{D_{KL}}\big[q_{\Phi}(\vec{z}_h; \vec{x}^{(n)})\,||\,p_{\Theta}(\vec{z}_h)\big]\biggl), 
    \label{eq:elbo}
\end{align}
where $\mathrm{D_{KL}}$ is the Kullback-Leibler divergence. 
\subsection{Computation of the Encoder Expectations}
We first compute the factor $q(\vec{s};\vec{z}, \vec{x}^{(n)})$ of the approximate posterior distribution in a numerically stable way via 
inner energies $E(\vec{x}, \vec{s}, \vec{z})$ \cite{drefs2022evolutionary}:
\begin{equation}
    E_\Theta(\vec{x}, \vec{s}, \vec{z}) = -\log(p_\Theta(\vec{x}, \vec{s}, \vec{z})),
\label{eq:energies}
\end{equation}
where $\log(p_\Theta(\vec{x}, \vec{s}, \vec{z}))$ are the log-joint probabilities.
The factor $q(\vec{s};\vec{z}, \vec{x}^{(n)})$ can be computed using standard Bayes' rule and the definition Eq.~\eqref{eq:energies} as follows:
\begin{equation}
    q(\vec{s};\vec{z}, \vec{x}^{(n)}) = \dfrac{\exp\Big(B^{(n)} - E_\Theta(\vec{x}^{(n)}, \vec{s}, \vec{z})\Big)}{\sum_{\vec{s}'}\exp\Big(B^{(n)} - E_\Theta(\vec{x}^{(n)}, \vec{s}', \vec{z})\Big)},
    \label{eq:exact_posterior}
\end{equation}
where $B^{(n)}= \underset{\vec{s}}{\min}\{E_\Theta(\vec{x}^{(n)}, \vec{s}, \vec{z})\}$ is used to ensure the exponentiation remains within the range of machine precision. The expectations of a function $f(\vec{x}, \vec{s}, \vec{z})$ w.r.t. factor of the approximate posterior $\mathbb{E}_{q(\vec{s};\vec{z}, \vec{x}^{(n)})}\big[f(\vec{x}^{(n)}, \vec{s}, \vec{z})\big]$ can then be computed as: %
\begin{align}
    \mathbb{E}_{q(\vec{s};\vec{z}, \vec{x}^{(n)})}&\big[f(\vec{x}^{(n)}, \vec{s}, \vec{z})\big] =\dfrac{\sum_{\vec{s}}f(\vec{x}^{(n)}, \vec{s}, \vec{z})\exp\Big(B^{(n)} - E_\Theta(\vec{x}^{(n)}, \vec{s}, \vec{z})\Big)}{\sum_{\vec{s}'}\exp\Big(B^{(n)} - E_\Theta(\vec{x}^{(n)}, \vec{s}', \vec{z})\Big)}.
    \label{eq:exp_wrt_posterior}
\end{align}
\subsection{Optimization of the Encoding and Decoding Models}
The ELBO of MS-VAE is now optimized w.r.t. the encoder's parameters $\Phi_{1:H}$ and decoder's parameters $W_{1:H}$ (further shortened as $\Phi$ and $W$, respectively), as well as the prior parameters $\vec{\pi}$ and the noise level $b$. We optimize the DNN parameters by approximating
gradients $\vec{\nabla}_{\Phi}\mathcal{F}(\Phi, W)$ and $\vec{\nabla}_{W}\mathcal{F}(\Phi, W)$. Concretely, we require for a given $\vec{x}^{(n)}$ a number of $M$ samples $\vec{z}_h^{\,(1:M)}$ for each $h$, i.e.,
\begin{equation}
    \vec{z}_h^{(m)} \sim q_{\Phi_h}(\vec{z}_h; \vec{x}^{(n)}).
    \label{eq:sampling}
\end{equation}
The approximate gradient for the encoder parameters $\Phi$ is then given by:
\begin{align}
    \vec{\nabla}_{\Phi}\mathcal{F}(\Phi, W) &\approx \vec{\nabla}_{\Phi}\Big[\dfrac{1}{NM^H}\sum_{n,m}\mathbb{E}_{q(\vec{s};\vec{z}^{(m)}, \vec{x}^{(n)})}\big[\log p_{\Theta}(\vec{x}^{(n)}|\vec{s},\vec{z}^{(m)})\nonumber \\
    &- \log q (\vec{s};\vec{z}^{(m)}, \vec{x}^{(n)})\big]
    - \textstyle \dfrac{1}{N}\sum_{n} \mathrm{D_{KL}}\big[q_{\Phi}(\vec{z}^{(m)};\vec{x}^{(n)})\,||\,p_{\Theta}(\vec{z}^{(m)})\big]\Big].
    \label{eq:nabla_phi}
\end{align}
To be able to compute gradients, we do not draw directly from the Gaussian distribution but use the reparameterization trick \cite{Kingma2014}: 
\begin{align}
   \vec{z}_h^{(m)} = \vec{\nu}_h(\vec{x}; \Phi_h) + \sqrt{\vec{\tau}_h(\vec{x};{\Phi_h})}\,\vec{\epsilon}_h^{(m)},\ \,\vec{\epsilon}_h^{(m)}\sim\mathcal{N}\big(\vec{\epsilon}_h; \vec{0}, \mathbb{I}\big).
   \label{eq:reparameterization_trick}
\end{align}

\begin{algorithm}[t]
\DontPrintSemicolon 
\caption{Training Multi-Stream Variational Autoencoders}
\SetKwInput{KwInit}{Initialize}
\KwInit{model parameters $\vec{\pi}, b$ and encoder parameters $\Phi$}
Initialize the MS-VAE model with $H$ discrete latent states\;
Load all pretrained decoder parameters $W$\;
\Repeat{parameters $\vec{\pi}, b$ and ELBO have converged}{
    \For{each batch in the dataset}{
        \For{each sample $\vec{x}^{(n)}$ in batch}{
            \For{each source $h = 1,\ldots,H$}{
                Encode: compute $\vec{\nu}_h^{(n)} = \vec{\nu}(\vec{x}^{(n)}; \Phi_h)$ and $\vec{\tau}_h^{(n)} = \mathrm{diag}(\vec{\tau}(\vec{x}^{(n)};\Phi_h))$ (see Eq.~\eqref{eq:encoder})\;
                \For{$m = 1,\ldots,M$}{
                    Sample and reparameterize $\vec{z}_h^{(m)}$ (Eqs.~\eqref{eq:sampling},\eqref{eq:reparameterization_trick})\;
                }
                Decode: compute $\vec{\mu}_h^{(m)} = \vec{\mu}(\vec{z}_h^{(m)}; W_h)$\;
            }
            Form combined encoder statistics: $\mathbf{\vec{\nu}}^{(n)}$, $\mathbf{\vec{\tau}}^{(n)}$\;
            Form combined decoder statistics: $\mathbf{z}^{(n)}$, $\vec{\mu}^{(n)}$\;
            Compute $\vec{\mu}_{\Theta_\mathrm{mix}}^{(m)}$ using Eq.~\eqref{eq:comb_model}\;
            Compute inner energies using Eq.~\eqref{eq:energies}\;
        }
        Compute batch ELBO using Eqs.~\eqref{eq:elbo},\eqref{eq:exp_wrt_posterior}\;
        Compute $\vec{\pi}_{\mathrm{batch}}$ and $b_{\mathrm{batch}}$ using Eq.~\eqref{eq:param-update}\;
        Update $\Phi$ using Adam (Eq.~\eqref{eq:nabla_phi})\;
        \If{decoder is not fixed}{
          Update $W$ using Adam (Eq.~\eqref{eq:nabla_W})\;
        }
        \Else{
          Skip decoder update\;
        }
    }
    Compute final $\vec{\pi}^{\mathrm{new}}$ and $b^{\mathrm{new}}$, ELBO as the mean over all batches\;
}
\label{alg:training_MSVAE}
\end{algorithm}

Similarly, for optimization of the decoder weights (when applicable), the gradient of the objective in Eq.~\eqref{eq:elbo} w.r.t. decoder's parameters $W$ can be approximated as:
\begin{equation}
  \vec{\nabla}_{W} \mathcal{F}(\Phi, W) \approx - \dfrac{1}{bNM^H} \sum_{n,m}\mathbb{E}_{q(\vec{s};\vec{z}^{(m)}, \vec{x}^{(n)})}\big[\vec{\nabla}_{W} |\vec{x}^{(n)} - \vec{\mu}^{(m)}_{\Theta_{\mathrm{mix}}}|\big]. 
  \label{eq:nabla_W}
\end{equation}
Here, we observe similarity to the standard VAE gradient ascent procedure, where automatic differentiation tools are applied to an L2 distance. However, in our case, it is an L1 distance due to Laplace distribution assumed for the observed data Eq.~\eqref{eq:gen_model_3}.

Finally, using the result in Eq.~\eqref{eq:exp_wrt_posterior}, the update equations for the remaining model parameters are derived by setting the corresponding derivatives of the ELBO to zero (no gradients required), these updates are given by:
\begin{align}
  \pi_h& \approx \dfrac{1}{NM^H}\sum_{n,\,m}\mathbb{E}_{q(\vec{s};\vec{z}^{(m)}, \vec{x}^{(n)})}\big[s_h\big], \nonumber \\
   b & \approx \dfrac{1}{DNM^H}\sum_{n,\,m}\mathbb{E}_{q(\vec{s};\vec{z}^{(m)}, \vec{x}^{(n)})}\big[|\vec{x}^{(n)}-\vec{\mu}^{(m)}_{\Theta_{\mathrm{mix}}}|\big].
  \label{eq:param-update}
\end{align}
The steps described in this section are summarized in Algorithm~\ref{alg:training_MSVAE}. Given a mixed signal, the encoder for source $h$ ideally infers the same $\vec{z}_h$ as it would without interference from other sources. In other words, the encoder must reliably extract information about its target source, regardless of how many other sources are active (see Fig.~\ref{fig:ms-vae-diagram}). If all encoders achieve this, signal reconstruction quality will be high, resulting in high ELBO values. Thus, optimizing the ELBO in Eq.~\eqref{eq:elbo} encourages the MS-VAE to encode distinct signal sources in separate streams. In contrast, each decoder’s task is to reconstruct its source from the latent state; if a decoder is already trained to reconstruct a source, minimal further optimization is needed, even in the presence of other sources. We leverage this property of MS-VAE when pretraining decoders with available labels.

\subsection{Discrete Case Prediction}
To determine which sources are present or absent in a given $\vec{x}^{(n)}$, we can use an approximation of $p(\vec{s}\,|\,\vec{x}^{(n)})=\int p(\vec{s},\vec{z}\,|\,\vec{x}^{(n)})\mathrm{d}\vec{z}$, where the marginalization over $\vec{z}$ is approximated again via sampling: 
\begin{equation}
   p(\vec{s}\,|\,\vec{x}^{(n)}) \approx q(\vec{s}; \vec{x}^{(n)}) \approx \dfrac{1}{M^H}\textstyle\sum_{m}q(\vec{s};\vec{z}_h^{(m)}, \vec{x}^{(n)}),
    \label{eq:q_s_x}
\end{equation}
with $q(\vec{s};\vec{z}, \vec{x}^{(n)})$ defined as in Eq.~\eqref{eq:exact_posterior}. 
It is also possible to find the factors of approximate posterior probability for individual sources $q(s_h=1;\vec{x}^{(n)})$:
\begin{equation}
    q(s_h = 1;\vec{x}^{(n)}) = \textstyle\sum_{\vec{s'}}q(\vec{s}';\vec{x}^{(n)})\delta(s'_h = 1).
    \label{eq:q_s_x_individual}
\end{equation}

\section{Experiments}
\label{sec:experiments}

The following subsections detail the individual numerical experiments. In all experiments, we used DNN architectures (encoders and decoders) from \cite{neri2021unsupervised}, each comprising five feed-forward linear blocks with ReLU activations and batch normalization. Input size 
$D$ and encoder layer dimensions are: $D=784$, (700-600-500-400-300) for MNIST, and $D=4369$, (428-320-224-160-80) for speech spectrograms. The continuous latent space has dimension $2\times Z\times H$, where $2$
stands for mean and variance in Eq.~\eqref{eq:encoder}, $Z=20$. Decoders are mirror architectures of the encoders. We used ADAM optimizer with negligible decay (0.02\% per epoch) and learning rates of $lr_1 = 0.0001$ (for all pretraining and Sec.~\ref{application_acoustic}) and $lr_2 = 0.0002$ (Sec.~\ref{proof_concept}-\ref{extension_K_sources}). The number of samples $M=100$ and batch size $B=8$ for Sec.~\ref{proof_concept} and standard $M=1$, $B=125$ for Sec.~\ref{extension_K_sources} and ~\ref{application_acoustic} due to increasingly large tensor dimensionality. All training and inference experiments were performed on an NVIDIA Tesla V100 GPU. The source code will be made publicly available together with a future accepted version of the paper.

\subsection{An Initial Proof of Concept}
\label{proof_concept}
\begin{figure}[h!]
    \centering
    \includegraphics[width=\linewidth]{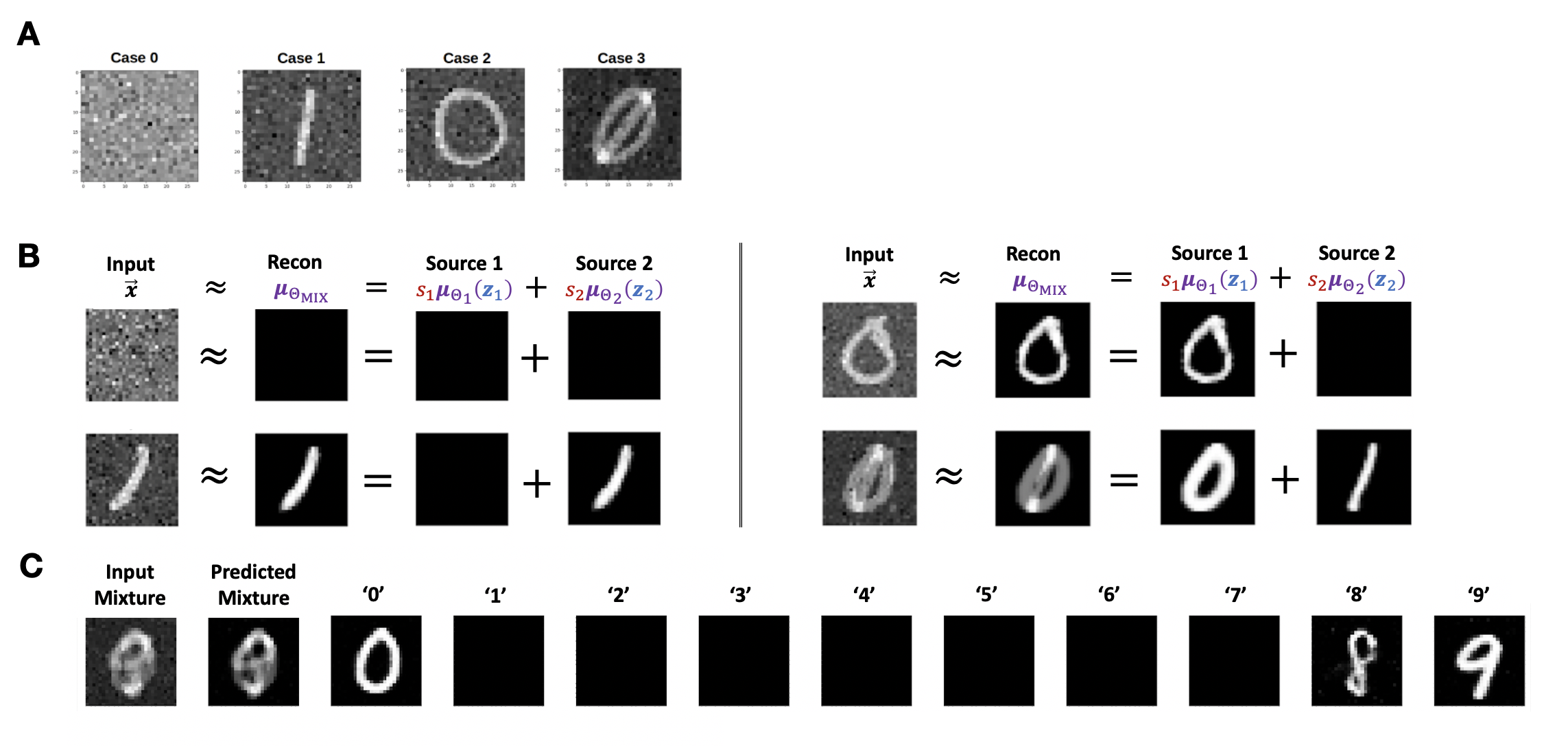}
    \caption{\textbf{A} Examples of generated artificial data points for subset of digits `0' and `1'. \textbf{B} Schematic of predicted results: mixtures and individual sources. \textbf{C}~Source disentanglement using the full MNIST dataset with 10\% of available labels; from left to right: input (a mixture of digits), MS-VAE reconstruction using different streams, individual reconstructions used by the ten individual VAE streams/experts (three streams were activated to reconstruct the here given input). Finetuning of the model leads to slightly better reconstruction, see Tab.~\ref{tab:mnist_results}.}
    \label{fig:schematic-results}
\end{figure}

For the initial proof of concept, we generated an artificial dataset using digits ‘0’ and ‘1’ from MNIST dataset \cite{lecun1998mnist}. Training and testing datasets ($N_{\mathrm{train}}=10^4$, $N_{\mathrm{test}}=10^3$) were generated according to the generative model with $H=2$ sources, $\pi_{1}^{\mathrm{gen}} = 0.3$, $\pi_{2}^{\mathrm{gen}} = 0.2$, $b^{\mathrm{gen}}=0.1$. For $H=2$, there are $2^H=4$ possible binary cases: no digits, one (either ‘0’ or ‘1’) or both digits (overlapping ‘0’ and ‘1’) are present, see Fig.~\ref{fig:schematic-results} \textbf{A}. We combine the sources linearly (assumption used in, e.g., \cite{jayaram2020source,neri2021unsupervised,postolache2023latent}) as given by the combination of sources model in Eq.~\eqref{eq:comb_model}. The parameters were initialized as $\pi_{1}^{\mathrm{init}} =\pi_{2}^{\mathrm{init}} = 0.5$, $b^{\mathrm{init}}=1$. The inference task for MS-VAE is to estimate, on previously unseen data, the activity of each source and to separate the digits when both are active. For the first stage of training, we pretrain two individual VAEs as “experts” trained exclusively on `0's or `1's, respectively, using all available digit class images in the MNIST training set. The expert networks are pretrained for $10^3$ epochs. After pretraining, the decoder weights are loaded into individual decoders of MS-VAE model, while the encoder linear layers are initialized with PyTorch’s default Kaiming/He initialization. The complete model is then trained for 100 epochs. Although the decoder weights could further be optimized during training, we keep them fixed as they already encapsulate the necessary information for digit reconstruction.
To assess MS-VAE performance, we first evaluated prediction accuracy and parameter convergence. After 100 epochs, the model achieved final a accuracy of $99.9\%$ with converged parameters $\pi_{1}^{\mathrm{100}} \approx 0.31$, $\pi_{2}^{\mathrm{100}} \approx 0.20$, and $b^{\mathrm{100}}\approx0.11$. We also conducted an individual entropy analysis and verified convergence of the ELBO to an entropy sum \cite{damm2023elbo} (applicable to MS-VAE as 
$b$ parameter is learned). For MS-VAE, the entropy sum can be 
computed as: 
\begin{align}
    \mathcal{H}_{\mathrm{sum}} &= \sum_{h=1}^{H}\Big( \dfrac{1}{N}\sum_{n=1}^{N}\mathcal{H}_{\mathrm{enc}}\big(q_{\Phi_h}(\vec{z}_h;\vec{x}^{(n)}\big)\Big) - H \, \mathcal{H}\big(p_\Theta(\vec{z})\big) - \mathcal{H}\big(p_\Theta(\vec{s})\big) \nonumber \\ 
    &+\dfrac{1}{N}\sum_{n=1}^{N}\mathbb{E}_{q_{\Phi}(\vec{z};\vec{x}^{(n)})}\mathcal{H}\big(q(\vec{s};\vec{z}, \vec{x}^{(n)})\big)-\dfrac{1}{N}\sum_{n=1}^{N}\mathcal{H}_{\mathrm{dec}}\big(p_\Theta\big(\vec{x}^{(n)}|\vec{s}, \vec{z}\big)\big).
    \label{eq:entropy_sum}
\end{align}
Evaluation confirmed convergence of ELBO to a entropy sum after 100 epochs $\mathcal{F}_{\mathrm{train}}^{\mathrm{100}}\approx412.0$, $\mathcal{H}_{\mathrm{sum}}^{\mathrm{100}} \approx 412.6$ and shown that the average posterior entropy $\mathcal{H}\big(q(\vec{s};\vec{x})\big)$ was approaching 0, indicating high posterior confidence\footnote{See \cite{damm2023elbo} for more details on entropy analysis}, coinciding with high prediction accuracy results. The qualitative digit prediction results are shown in Fig.~\ref{fig:schematic-results} \textbf{B}.

\subsection{Extension to multiple sources}
\label{extension_K_sources}
After confirming the model's viability, we extended the approach to all MNIST digits. 
For the extended experiment, we now used either 100\% or 10\% of the training data to pretrain the decoders of the ten experts (one expert per digit). We denoted a MS-VAE pretrained with all labels by MS-VAE (100\%) and MS-VAE (10\%) if merely 10\% of the labels were used.
Decoders pretrained on 100\% of the data were used as in Sec.~\ref{proof_concept}. For the MS-VAE (10\%) , decoders were loaded and fixed during initial 100 epochs, then included in the final computational graph and further finetuned for additional 50 epochs. We generated training and testing datasets ($N_{\mathrm{train}}=5\times10^4$, $N_{\mathrm{test}}=5\times10^3$), including all $H=10$ MNIST digits, each having generative Bernoulli prior of $\pi_{h}^{\mathrm{gen}} = 1/H = 0.1$, and generative noise $b^{\mathrm{gen}}=0.1$. The resulting number of binary cases is $2^{H} =1024$. The parameters were initialized as in Sec.~\ref{proof_concept}. We observed convergence of prior parameters in both training setups: $\pi_{h}^{\mathrm{100}}=\pi_{h}^{\mathrm{150}} \approx 0.1$, $b^{\mathrm{100}} = b^{\mathrm{150}} \approx 0.11$ and the ELBO convergence to entropy sums: $\mathcal{F}_{\mathrm{train}}^{\mathrm{100}}\approx\mathcal{H}_{\mathrm{sum}}^{\mathrm{100}} \approx 340.0$ (MS-VAE 100\%),  $\mathcal{F}_{\mathrm{train}}^{\mathrm{150}}\approx\mathcal{H}_{\mathrm{sum}}^{\mathrm{150}} \approx 354.0$ (MS-VAE 10\%). We also observed high average prediction accuracies (over 3 test runs): 91.23±0.14\% (MS-VAE 100\%), 90.80±0.28\% (MS-VAE 10\%) and $\mathcal{H}\big(q(\vec{s};\vec{x})\big) \rightarrow 0$. 
\begin{table}[h!]
\caption{Quantitative comparison of average source separation metrics with population standard deviation (3 inference runs). Higher PSNR/SSIM values are better.}
\label{tab:mnist_results}
\begin{adjustbox}{max width=\columnwidth}
\begin{tabular}{lcccccccc}
\hline
\multicolumn{1}{c}{}                                       
& \begin{tabular}[c]{@{}c@{}}\textbf{Inference}\\ \textbf{time, s}\end{tabular} 
& \begin{tabular}[c]{@{}c@{}}\textbf{PSNR}\\ \textbf{K=2, dB}\end{tabular} 
& \begin{tabular}[c]{@{}c@{}}\textbf{SSIM}\\ \textbf{K=2}\end{tabular} 
& \begin{tabular}[c]{@{}c@{}}\textbf{PSNR}\\ \textbf{K=3, dB}\end{tabular} 
& \begin{tabular}[c]{@{}c@{}}\textbf{SSIM}\\ \textbf{K=3}\end{tabular} & \begin{tabular}[c]{@{}c@{}}\textbf{PSNR}\\ \textbf{K=4, dB} \end{tabular} & \begin{tabular}[c]{@{}c@{}}\textbf{SSIM}\\ \textbf{K=4}\end{tabular}\\ 
\hline
\begin{tabular}[c]{@{}l@{}}
Average\end{tabular} & - & 14.75{\color[HTML]{9B9B9B}\,±0.01} & 0.61{\color[HTML]{9B9B9B}\,±0.00} & 13.48{\color[HTML]{9B9B9B}\,±0.00} & 0.47{\color[HTML]{9B9B9B}\,±0.00} & 12.98{\color[HTML]{9B9B9B}\,±0.00} & 0.39{\color[HTML]{9B9B9B}\,±0.00} \\
\hline

VAE-BSS & 0.0017{\color[HTML]{9B9B9B}\,±0.0003} & 13.92{\color[HTML]{9B9B9B}\,±0.00} & 0.54{\color[HTML]{9B9B9B}\,±0.00} & 11.97{\color[HTML]{9B9B9B}\,±0.01} & 0.39{\color[HTML]{9B9B9B}\,±0.00} & 10.94{\color[HTML]{9B9B9B}\,±0.00} & 0.30{\color[HTML]{9B9B9B}\,±0.00}\\

VAEM-BSS & 0.0019{\color[HTML]{9B9B9B}\,±0.0006}                   & 17.01{\color[HTML]{9B9B9B}\,±0.02} & 0.75{\color[HTML]{9B9B9B}\,±0.00} & 12.68{\color[HTML]{9B9B9B}\,±0.01} & 0.47{\color[HTML]{9B9B9B}\,±0.00} & 11.50{\color[HTML]{9B9B9B}\,±0.00} & 0.37{\color[HTML]{9B9B9B}\,±0.00}\\ 
\hline

LASS (top-k, k=32)& 0.0065{\color[HTML]{9B9B9B}\,±0.0007} & 13.21{\color[HTML]{9B9B9B}\,±0.01} &0.60{\color[HTML]{9B9B9B}\,±0.00} & -$^*$ & -$^*$ & -$^*$ & -$^*$ \\
\hline

BASIS (NCSN) & 23.801{\color[HTML]{9B9B9B}\,±3.3956} & 30.23{\color[HTML]{9B9B9B}\,±0.02} &0.92{\color[HTML]{9B9B9B}\,±0.00} & 14.03{\color[HTML]{9B9B9B}\,±0.02} & 0.49{\color[HTML]{9B9B9B}\,±0.00} & -$^\diamond$  & -$^\diamond$ \\
\hline

\textbf{MS-VAE} (100\%) & 0.0216{\color[HTML]{9B9B9B}\,±0.0001} & 16.76{\color[HTML]{9B9B9B}\,±0.03} & 0.76{\color[HTML]{9B9B9B}\,±0.00} & 14.96{\color[HTML]{9B9B9B}\,±0.03} & 0.66{\color[HTML]{9B9B9B}\,±0.00} & 13.64{\color[HTML]{9B9B9B}\,±0.02} & 0.58{\color[HTML]{9B9B9B}\,±0.00}\\

\begin{tabular}[c]{@{}l@{}}\textbf{MS-VAE} (10\%)\end{tabular} & 0.0217{\color[HTML]{9B9B9B}\,±0.0001} & 16.87{\color[HTML]{9B9B9B}\,±0.02} & 0.72{\color[HTML]{9B9B9B}\,±0.00} & 15.20{\color[HTML]{9B9B9B}\,±0.03} & 0.63{\color[HTML]{9B9B9B}\,±0.00} & 13.98{\color[HTML]{9B9B9B}\,±0.02} & 0.56{\color[HTML]{9B9B9B}\,±0.00}\\ 
\hline
\multicolumn{4}{l}{\small $^*$ -- LASS is applicable only for $K=2$} \\
\multicolumn{4}{l}{\small $^\diamond$ -- NaN output values observed} \\
\end{tabular}
\end{adjustbox}
\end{table}

Although MS-VAE is an application agnostic model, it can be applied to a standard task of source separation. For our comparison, we chose readily available pretrained models of an unsupervised VAE for (Masked) Blind Source Separation (VAE(M)-BSS) model \cite{neri2021unsupervised} for $K=2,\,3, 4$ and two Bayesian approaches which leverage pretrained neural network priors, namely: Latent Autoregressive Source Separation (LASS) \cite{postolache2023latent} (only applicable for $K=2$) and Bayesian Annealed Signal Source Separation (BASIS) \cite{jayaram2020source} (with a Noise Conditioned Score Network (NCSN) \cite{song2019generative} prior). 

All the approaches pursue different training strategies. In training of VAE(M)-BSS, each epochs involves random pairing of images from the MNIST training set ($6 \times 10^4$ data points) and splitting the mini-batches in half to produce final mixed inputs. Although the theoretical maximum size of mixtures in the training set ($K=2$) is given by the binomial coefficient $\binom{6 \times 10^4}{2} \approx 1.8 \times 10^9$ (as also noted by authors), in reality, only $\approx3 \times 10^4$ mixtures are generated per epoch, making the cumulative number of unique mixtures $\approx 5000 \times 3 \times 10^4 \approx 1.5 \times 10^8$ over $5000$ epochs, which is still, however, a considerable training set size. VAE(M)-BSS produces reconstructions of individual digit as well as mixed digits. Additional $\beta$ annealing is applied to $\mathrm{D_{KL}}$ term to prevent early posterior collapse. For inference, optionally, the sum of estimated sources is constrained to match the mixture (``masking''), improving the quality of reconstructed digits. We further refer to the approach with and without masking as VAEM-BSS and VAE-BSS, respectively. LASS employs a two-stage pretraining strategy: a VQ-VAE is first trained on individual MNIST digits to produce compact discrete latent codes, which are subsequently used to train an autoregressive Transformer which predicts the probability of a given digit and serves as a prior for inference. BASIS assumes Gaussian-distributed mixtures $\vec{m}\sim \mathcal{N}(g(\vec{x}), \gamma^2I)$, where $g(\vec{x})$ is a mixture image (linearly superimposed digits). The MNIST training set is split into 48000 train and 12000 test data points. NCSN is then trained for $2\times 10^5$ epochs using denoising score matching loss to match the derivative of the log-density (i.e., the score function) of the data distribution. Similarly to LASS, the pretrained NCSN is then finally used as a deep generative prior during inference. 

The diverging training procedures described above highlight the difficulty in comparing different approaches. On the one hand, it is possible to evaluate the methods on the same, previously established task; specifically, on separation of a fixed number $K$ of linearly superimposed digits. On the other hand, the approaches capable of addressing this task, including our own, employ different training protocols. Therefore, our comparison must account for differences in training procedures and available information across methods. Where relevant, we note these differences in our comparisons. 

\begin{figure*}[t]
    \centering
    \includegraphics[width=\linewidth]{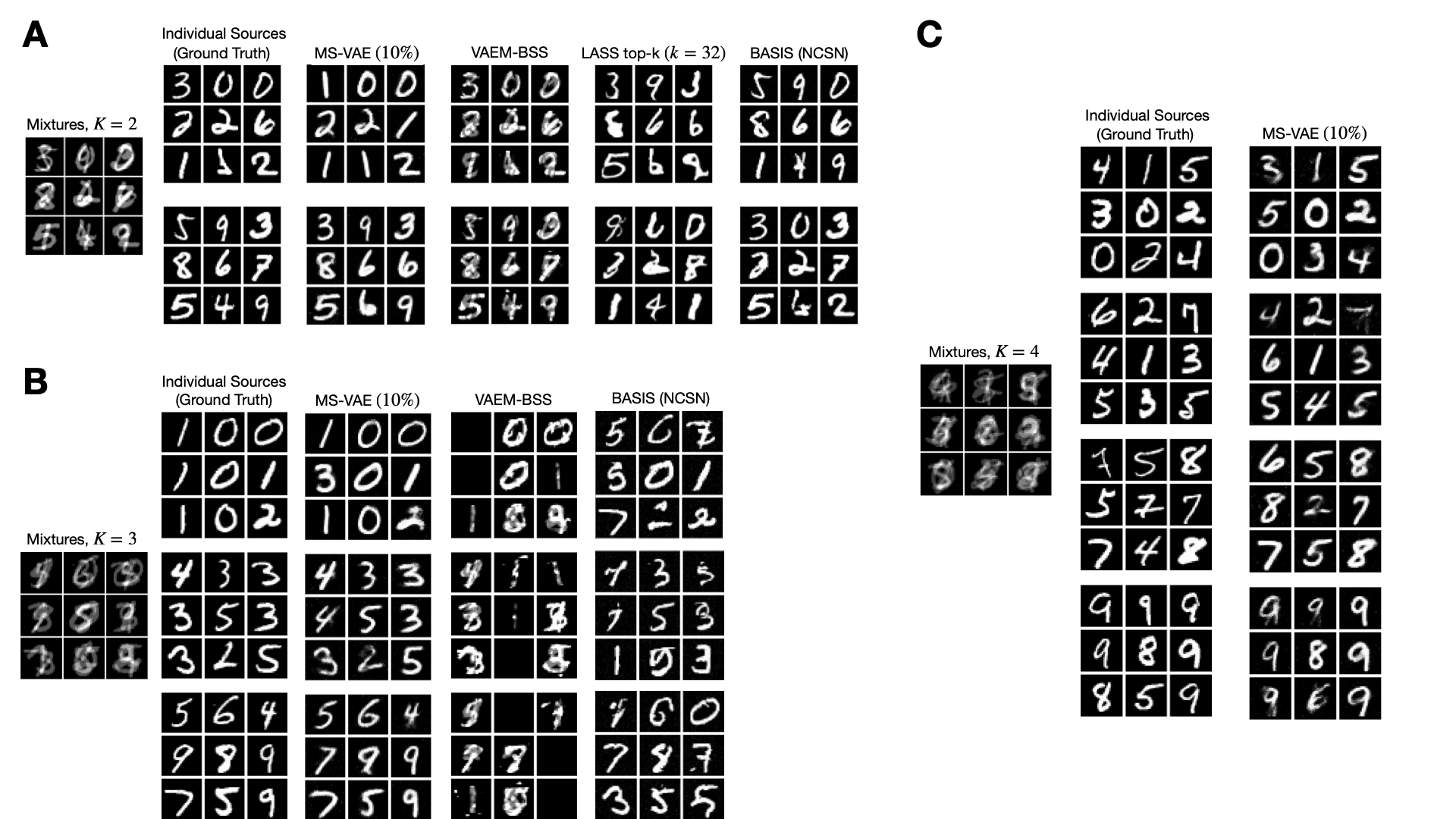}
    \caption{Examples of source separation performance of MS-VAE (10\%), VAEM-BSS, LASS and BASIS (NCSN) models for \textbf{A} $K=2$ sources, \textbf{B} $K=3$ sources and \textbf{C} $K=4$ sources. The permutation of sources can sometimes be observed in the outputs of, e.g., MS-VAE, LASS and BASIS.}
    \label{fig:mnist-result-K-sources}
\end{figure*}

To address the source separation task, we use the same MS-VAE (100\%) and MS-VAE (10\%) models trained with a Bernoulli prior which we described in the beginning of the section (no retraining was required).
The model can be adapted by simply replacing the Bernoulli prior (see Eq.~(\ref{eq:gen_model_1}) with fixed-K prior formulation appropriate for this task, while leaving all weights and biases of the individual encoder-decoder pairs unchanged:
\begin{align}
    p_\Theta (\vec{s}) = \begin{cases}
        \binom{H}{K}^{-1} &\mathrm{if} \, \sum_{h=1}^{H}s_h = K \\
        \,0, &\mathrm{otherwise}, 
    \label{eq:fixed_K_prior}
\end{cases}
\end{align}
where $\binom{H}{K}^{-1}$ is the reciprocal binomial coefficient. $H$, as previously defined, is total number of sources $H=10$.
The computations of the factor $q(\vec{s};\vec{z}, \vec{x}^{(n)})$ can then be simplified as follows (compare to Eq.~(\ref{eq:exact_posterior})): 
\begin{equation}
    q(\vec{s};\vec{z}, \vec{x}^{(n)}) = \dfrac{\exp\Big(B^{(n)} - E_\Theta(\vec{x}^{(n)}, \vec{s}, \vec{z})\Big)\delta(\vec{s}\in\mathcal{R}_K)}{\sum_{\vec{s}'\in\mathcal{R}_K}\exp\Big(B^{(n)} - E_\Theta(\vec{x}^{(n)}, \vec{s}', \vec{z})\Big)},
    \label{eq:exact_posterior_fixed_K}
\end{equation}
where $\mathcal{R}_K \coloneq \{\vec{s}\in\{0, 1\}^H|\sum_{h=1}^{H}s_h = K\}$. The computations of expectations w.r.t. $q(\vec{s};\vec{z}, \vec{x}^{(n)})$ can be simplified to (compare to Eq.~(\ref{eq:exp_wrt_posterior})):
\begin{align}
    \mathbb{E}_{q(\vec{s};\vec{z}, \vec{x}^{(n)})}&\big[f(\vec{x}^{(n)}, \vec{s}, \vec{z})\big] =\dfrac{\sum_{\vec{s}\in\mathcal{R}_K}f(\vec{x}^{(n)}, \vec{s}, \vec{z})\exp\Big(B^{(n)} - E_\Theta(\vec{x}^{(n)}, \vec{s}, \vec{z})\Big)}{\sum_{\vec{s}'\in\mathcal{R}_K}\exp\Big(B^{(n)} - E_\Theta(\vec{x}^{(n)}, \vec{s}', \vec{z})\Big)}.
    \label{eq:exp_wrt_posterior_fixed_K}
\end{align}
This means that, rather than evaluating all $1024$ binary cases, the number is reduced to 45, 120 and 210 for K=2, 3 and 4, respectively.
Results on test images always containing $K=2,3,4$ overlapping sources in terms of peak signal-to-noise ratio (PSNR) and structural similarity index (SSIM) can be seen in Tab.~\ref{tab:mnist_results}. During evaluation it was observed that output sources can be permuted, hence, we compute the objective metrics in a permutation-invariant way. For $K=2$, BASIS (NCSN), despite having the longest inference times, maintains its state-of-the-art performance in terms of PSNR, followed by VAEM-BSS (see Fig.~\ref{fig:mnist-result-K-sources} \textbf{A} for qualitative comparison). However, in the more challenging setting $K=3$, MS-VAE is able to improve the performance in both objective metrics and qualitative separation (see Fig.~\ref{fig:mnist-result-K-sources} \textbf{B}). The different MS-VAE versions also surpass the Average and VAEM-BSS for $K=4$, with qualitative separation results of MS-VAE 10\% shown in Fig.~\ref{fig:mnist-result-K-sources} \textbf{C}.

As for the disentanglement performance, mean classifier accuracy based on Eq.~\eqref{eq:q_s_x} can be viewed as a disentanglement measurement similar to the Z-min score \cite{carbonneau2022measuring}. Although a direct comparison to the unsupervised JointVAE \cite{dupont2018learning} is not entirely fair (due to the same reasons as with other source separation approaches), we include it here, as, to our knowledge, it is the only VAE-based disentanglement model incorporating continuous and discrete features which reports an accuracy of 88.7\% for digit classification on the full MNIST dataset. JointVAE was trained and tested on non-overlapping digits (one digit per image), hence, we tested MS-VAE models on images containing $K=1$ sources.  
In this scenario, the average classifier accuracies over three independent runs are 95.82±0.24\% for MS-VAE (100\%) and 94.72±0.30\% for MS-VAE (10\%). 

\subsection{Application to Acoustic Data}
\label{application_acoustic}
Speaker diarization is the most semantically related acoustic application, in which each discrete latent $s_h$ corresponds to the activity of a specific speaker.
To test our approach, we used a $\SI{358}{\second}$ recording of a conversation (2 males, 1 female) in quiet, recorded at $fs=\SI{48}{\kilo\hertz}$, from a recent conversational behavior database \cite{hinrichs2025database}.
Due to a mismatch between the estimated statistics for one of the male speakers (specifically, the parameters $\pi_{h}^{\mathrm{train}}$ and $\pi_{h}^{\mathrm{test}}$ were significantly different so that reasonable application of our approach was not possible), we retained the recordings of the other male (speaker 1) and the female (speaker 2) participants. The estimated label-based statistics: $\pi_{1}^{\mathrm{train}} \approx 0.30$, $\pi_{1}^{\mathrm{test}} \approx 0.27$; $\pi_{2}^{\mathrm{train}} \approx 0.37$, $\pi_{2}^{\mathrm{test}} \approx 0.36$. The close-microphone recordings were mixed and segmented into overlapping frames of length $T\approx\SI{100}{\milli\second}$, then STFT transformed (rectangular window, 256 hop), yielding 256 frequency $\times$ 17 time bins per frame. Magnitude spectrograms were also normalized to [0, 1]. The first 120 seconds ($N_{\mathrm{test}}=2812$) were used for testing and the remaining 238 seconds ($N_{\mathrm{train}}=5602$) for training. Individual training recordings from close-microphones were processed similarly, with either 100\% or 10\% of data used for expert pretraining. All MS-VAE models were trained for 150 epochs with fixed decoders (in contrast to Sec.~\ref{extension_K_sources}).

In the close-microphone recordings, the presence of speech from other participants was detected. Applying a pretrained competitive Pyannote Diarization V2.1 model \cite{Bredin2021,Bredin2020} to the mixture of recordings directly produced a DER of 96.73\% ($K=2$ was specified to the model). To address this issue, we replaced the inactive frames with zeros (for Pyannote Diarization V2.1) or $x_{\mathrm{train/test}}\sim\mathrm{Laplace}(\vec{0}, 0.0008)$ (for MS-VAE).

\begin{figure}[h!]
    \centering
    \includegraphics[width=\columnwidth]{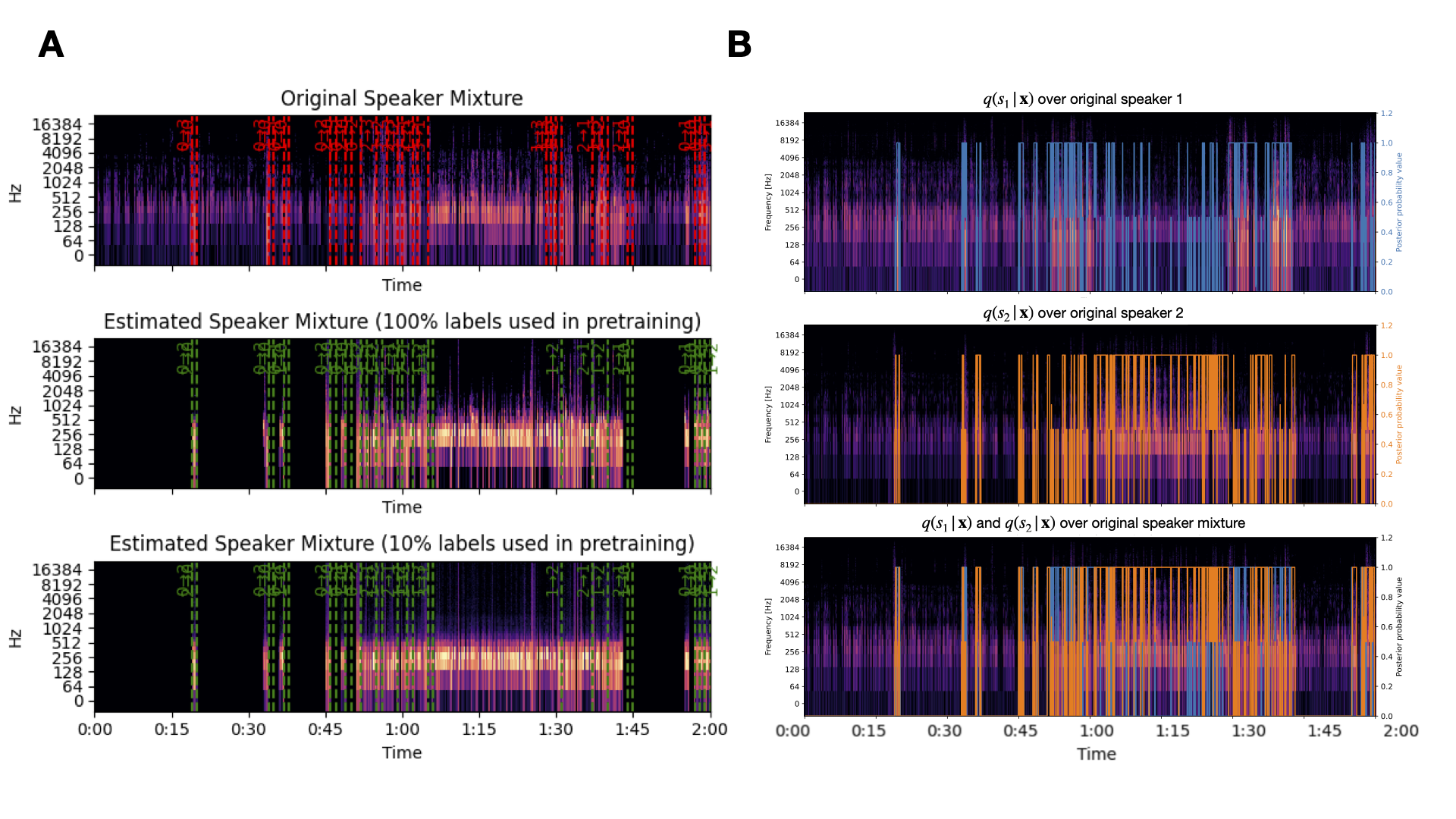}
    \caption{\textbf{A} Spectrograms of original/estimated mixtures with decision boundaries indicating case changes, for example, $0 \rightarrow 2$: silence to speaker 1, etc. Red: ground truth; green: predicted (Eq.~\ref{eq:q_s_x}) averaged over $\SI{1}{\second}$ intervals. \textbf{B} Plots of individual approximate factors of the posteriors (Eq.~\ref{eq:q_s_x_individual}) over spectrograms of original signals. Top: for speaker 1, middle: for speaker 2, bottom: for the conversation (mixture). The individual speaker's activation thresholds $\lambda_k$ can be calculated using these posterior probabilities and parameters $\pi_k$, so that: $A_k/C \approx \pi_k$, where $A_k$ is the area under the curve of $k$th source (controlled by $\lambda_k$) and $C$ is the total area. In our experiments, we used $\lambda_1 = \lambda_2 = 0.99$, as we observed overall best diarization performance.}
    \label{fig:diar-results}
\end{figure}
\begin{table}[h!]
\caption{Quantitative comparison in speaker diarization, with population standard deviation. MS-VAE diarization time annotations were derived from individual factors of posterior distribution Eq.~\eqref{eq:q_s_x_individual}, predictions averaged over $\SI{1}{\second}$ intervals.}
\label{tab:diar_results}
\begin{adjustbox}{max width=\columnwidth}
\begin{tabular}{lccccccc}
\hline
& \multicolumn{1}{c}{\textbf{\begin{tabular}[c]{@{}c@{}}Inference \\ time,\,s\end{tabular}}} & \multicolumn{1}{c}{\textbf{\begin{tabular}[c]{@{}c@{}}Average frame-\\wise accuracy,\,\%\end{tabular}}} & \multicolumn{1}{c}{\textbf{DER,\,\%}}     & \multicolumn{1}{c}{\textbf{FA,\,\%}}     & \multicolumn{1}{c}{\textbf{Miss,\,\%}}  & \multicolumn{1}{c}{\textbf{Conf,\,\%}}  & \multicolumn{1}{c}{\textbf{Corr.,\,\%}} \\ 
\hline
\begin{tabular}[c]{@{}l@{}}\textbf{MS-VAE} \\(100\%)\end{tabular} & 13.16{\color[HTML]{9B9B9B}\,±0.16} & 81.57{\color[HTML]{9B9B9B}\,±0.11}      & 29.90{\color[HTML]{9B9B9B}\,±0.69} & 22.55{\color[HTML]{9B9B9B}\,±0.0}     & 2.94{\color[HTML]{9B9B9B}\,±0.0} & 4.41{\color[HTML]{9B9B9B}\,±0.0}        & 92.65{\color[HTML]{9B9B9B}\,±0.0} \\

\begin{tabular}[c]{@{}l@{}}\textbf{MS-VAE} \\(10\%)\end{tabular} & 12.70{\color[HTML]{9B9B9B}\,±0.23} & 78.37{\color[HTML]{9B9B9B}\,±0.03}      & 38.24{\color[HTML]{9B9B9B}\,±0.0} & 29.41{\color[HTML]{9B9B9B}\,±0.0}      & 4.41{\color[HTML]{9B9B9B}\,±0.0} & 4.41{\color[HTML]{9B9B9B}\,±0.0}        & 91.18{\color[HTML]{9B9B9B}\,±0.0} \\ 
\hline
\begin{tabular}[c]{@{}l@{}}\textbf{Pyannote}\\ Diarization V2.1\end{tabular} & 46.04{\color[HTML]{9B9B9B}\,±0.50} & \multicolumn{1}{c}{-}                 & 18.36{\color[HTML]{9B9B9B}\,±0.0}  & 0.51{\color[HTML]{9B9B9B}\,±0.0} & 17.54{\color[HTML]{9B9B9B}\,±0.0} & 0.31{\color[HTML]{9B9B9B}\,±0.0} & 82.14{\color[HTML]{9B9B9B}\,±0.0}\\ 
\hline
\end{tabular}
\end{adjustbox}
\end{table}
Results on the preprocessed test audio, averaged over three inference runs, are shown in Tab.~\ref{tab:diar_results}. While MS-VAE models exhibit higher DER than Pyannote, they yield approximately $15\%$ less missed speech (Miss.) and about $11\%$ more precise speaker attribution (Corr.). These qualities might be especially desirable in, e.g., forensics to detect and attribute the perpetrators' speech if voice samples are available \cite{nagavi2024comprehensive} as in our experimental setup. In addition, fewer intervals were predicted by Pyannote (13 out of 16), which may contribute to lower DER. Notably, pretraining the decoder with just 10\% of the data produces similar decision boundaries, as shown in Fig.~\ref{fig:diar-results} \textbf{A}. Finally, we also remark that MS-VAE models have access to individual approximate factors of the posterior. They can thus be optionally used to tune the thresholds $\lambda_h$ and, hence, to control Miss-FA tradeoff (see Fig.~\ref{fig:diar-results} \textbf{B}).

\section{Discussion}
\label{sec:discussion}
In this paper, we introduced and explored the novel concept of a multi-stream VAE (MS-VAE). The approach incorporates source disentanglement by design. Our approach is application-agnostic and demonstrating a proof-of-concept was our primary goal, and no large deep neural networks (DNNs), or intricate DNNs with task/domain optimized architectures were used. Also no external training on datasets was used.

Nevertheless, we observed improvements over baseline methods in the disentanglement of mixed images (Tab.~\ref{tab:mnist_results}) and in speaker diarization (Tab.~\ref{tab:diar_results}). 
In disentanglement of images, examples like those in Fig.~\ref{fig:schematic-results} \textbf{C} demonstrate highly interpretable and clear-cut separation of individual sources (digits `0',`8' and `9'), with MS-VAE providing separate reconstructions for each digit. We also remark the robustness of performance of MS-VAE models: even for complex mixtures of sources (e.g., $K=3,4$ mixed sources), source separation maintains a high quality. In speaker diarization (Tab.~\ref{tab:diar_results}), we observed significantly fewer missing speech frames compared to \cite{Bredin2021,Bredin2020}, even though the latter is trained on large data corpora. In contrast, the MS-VAE approach requires significantly smaller amount of training data. Some supervision is required for the MS-VAE model, hence, it is not fully unsupervised like \cite{dupont2018learning} or \cite{neri2021unsupervised}. However, recent research suggests that some level of supervision may generally be necessary for effective disentanglement \cite{locatello2020sober}. For real data (Sec.~\ref{application_acoustic}), it is important that training and testing sets have similar statistics for successful model learning — a requirement not shared by methods such as \cite{Bredin2020}. MS-VAE also tends to hypothesize more speech segments when compared to the ground truth.

The main goal of this paper was to contribute to the largely unsolved challenge of learning disentangled representations. Our decisive novelty is the use of individual experts (VAE sub-networks) for each source. These experts are combined via discrete latents using an explicit source combination model (see Eq.~\ref{eq:comb_model}). Future work will explore additional explicit models, such as masking \cite{mousavi2023generic} and occlusion \cite{LuckeEtAl2009}) and more expressive priors. While combining larger numbers of sources increases the computational complexity, up to ten sources could be readily handled with current computing resources. More sources could be managed in the future using variational approximations for the discrete latents. Recent contributions demonstrated that generative models with discrete latents can be trained very efficiently \cite{boukun2024blind,drefs2022evolutionary,drefs2023direct,HirschbergerEtAl2022,mousavi2023generic}, suggesting such extensions will not introduce significant computational bottlenecks. Moreover, discrete encoding may in general be important for achieving high-quality disentanglement representations.

\section{Acknowledgments} 
We thank Jan Warnken for his help on derivations. This work was supported by the German Research Foundation (DFG) under grant ID 352015383 (SFB 1330, B2). We also thank the fellow researchers of the same collaborative grant (Paula Hinrichs, Volker Hohmann, Giso Grimm, project B1), who provided the conversational behavior database.

\bibliographystyle{IEEEbib}
\def\authornoop#1{}

\end{document}